\documentclass{article}

\usepackage[final]{neurips_2023}

\usepackage[utf8]{inputenc} 
\usepackage[T1]{fontenc}    
\usepackage{hyperref}       
\usepackage{url}            
\usepackage{booktabs}       
\usepackage{amsfonts}       
\usepackage{nicefrac}       
\usepackage{microtype}      
\usepackage{xcolor}         
\usepackage{graphicx}
\usepackage{wrapfig}

\title{Towards Ethical Multimodal Systems}

\author{%
  Alexis Roger \\
  Department of Computer Science and Operations Research \\
  University of Montreal\\
  Mila - Quebec AI Institute \\
  \texttt{alexis.roger@umontreal.ca} \\
  \And
  Esma Aïmeur \\
  Department of Computer Science and Operations Research \\
  University of Montreal\\
  \And
  Irina Rish \\
  Department of Computer Science and Operations Research \\
  University of Montreal\\
  Mila - Quebec AI Institute \\
}

\begin{document}

\maketitle

\begin{abstract}
  Generative AI systems (ChatGPT, DALL-E, etc) are  expanding into multiple areas of our lives, from art \cite{stableDiffusion} to mental health \cite{koko};  their rapidly growing societal impact opens new opportunities, but also raises ethical concerns. The emerging field of AI alignment aims to make AI systems reflect human values. This paper focuses on evaluating the ethics of multimodal AI systems involving both text and images - a relatively under-explored area, as most alignment work is currently  focused  on language models. We first create a multimodal ethical database from human feedback on ethicality. Then, using this database, we develop algorithms, including a RoBERTa-large classifier and a multilayer perceptron, to automatically assess the ethicality of system responses.
\vspace{-0.2in}
\end{abstract}

\section{Introduction} \label{sec:intro}
\vspace{-0.1in}
AI ethics encompasses safety, security, human concerns, and environmental issues, gaining significance in high-school education and AI literacy, as highlighted, for example, in \cite{forbes}. Addressing AI ethics is critical, underscored in papers like {\it Concrete Problems in AI Safety} (\cite{concreteProbs}) and {\it Unsolved Problems in ML Safety} (\cite{unsolvedProbs}).

Efforts to regulate AI ethics include the {\it Montreal Declaration for a Responsible Development of Artificial Intelligence} (\cite{decResponsible}). The United States is developing an {\it AI Bill of Rights} (\cite{aibor}), and Europe is working on {\it The Artificial Intelligence Act} (\cite{euai}), all aimed at ensuring AI's safe development. Projects like the Moral Machine by \cite{moralmachine} address AI ethical concerns. IBM's {\it AI Fairness 360} toolkit (\cite{fairness360}) offers tools to detect bias and promote fairness in algorithms, addressing issues like age or gender discrimination. However, none of these initiatives have focused on large multimodal models.

\noindent{\bf Our Contributions.} Similar to the Moral Machine project (\cite{moralmachine}), our paper aims to create a dataset of ethical and unethical samples. However, our focus is on multimodal models, which take both an image and a question as input and provide an answer. In this paper, we present our results for the MAGMA model by \cite{magma}, while evaluation of other similar multimodal models remains the topic of our ongoing work. Note that, as shown in Figure \ref{fig:pretest}, MAGMA can sometimes generate responses not well-aligned with common ethical standard, thus serving as a motivation for our line of work.

Our first contribution is a dataset consisting of 789 question and image pairs covering various ethical fields, including economics, medicine, society and discrimination. Each pair includes an answer generated by the MAGMA model by \cite{magma}, along with user votes indicating whether it's ethical, unethical, or unclear. Unique user IDs are included to track responses and identify abnormal behavior. This dataset, detailed in section \ref{sec:data}, maintains a balanced proportion of ethical and unethical responses.

Our second contribution is the development of an ethical evaluation pipeline via crowd-sourcing. This lightweight, user-friendly pipeline gamifies the evaluation process and is designed around Discord, a popular messaging service. Although initially used for multimodal prompt evaluations, it can be applied to other evaluation processes requiring more than a simple "like" from social media, as explained in section \ref{sec:data}.

Our third contribution involves a preliminary study of classification algorithms to automatically assess multimodal systems. We evaluate existing models and build our own, based on a multilayer perceptron using text and image embeddings as input. We delve into this comparison in section \ref{sec:classification}.
\vspace{-0.1in}

\section{Related Work} \label{sec:related}
\vspace{-0.1in}
\noindent{\bf State of the art.} Ethical considerations regarding Natural Language Processing (NLP) models have been explored in prior research, notably in \cite{roberta}. This study assessed various NLP models, including GPT-3 (few-shot learner) (\cite{model:gpt}), BERT (\cite{model:bert}), RoBERTa-large (\cite{model:roberta}), word averaging (\cite{model:average}) and ALBERT-xxlarge (\cite{model:albert}), across five key values: justice, deontology, virtue, utilitarianism, and commonsense. Larger fine-tuned models trained on more data generally exhibited better performance. However, these models primarily operated with textual input, while our focus is on multimodal input, combining text and images.

A study on the ethics of multimodal systems was conducted by \cite{mais}, which involved evaluating the ethics of a chosen multimodal algorithm and exploring avenues for model improvement. Notably, the authors faced limitations due to a small hand-crafted dataset and manual evaluation, which introduced bias and hindered scalability. Our aim is to address these challenges.

\noindent{\bf Background.}
In this study we focus on a text and image combination as input with a text output. Herein, we used the  MAGMA model by \cite{magma}, as the authors made both the code (\cite{MAGMAgit}) and a checkpoint of the trained model publicly available; evaluating more recent and more advanced models of this type is a topic of our ongoing work.  MAGMA is based on the CLIP visual encoder (\cite{clip}) and the GPT-J language model (\cite{gptJ}). MAGMA is also the model on which the previous paper from \cite{mais} was based on. In that paper the authors showed MAGMA had some critical ethical deficiencies (e.g., see Figure \ref{fig:pretest}). We therefore decided to expand on this analysis.

In order to perform such an ethical evaluation, we needed a reliable dataset of ethical and non-ethical examples on which we could rely. Due to the multimodal aspect of our study, we were unable to find any available datasets. In the preliminary study, \cite{mais} had used a small hand-crafted dataset as a proof of concept. This database contained a few tens of examples crafted and evaluated by the authors. This would not be scalable as we were hoping to build a dataset of about a thousand examples. To this end we looked at the paper by \cite{interactive} where they present a technique they used to gather human feedback. After reviewing their method, we realized that we needed to have a clear and straight-forward interface. Along with this we also needed a way to evaluate the participating people, to ensure they had proper ethics. These became our design goals for our own feedback gathering pipeline.


\section{Crafting an Ethical Dataset} \label{sec:data}
\vspace{-0.1in}
In order to build an ethical database we considered different alternatives. Many constraints had to be taken into consideration when building this database. For example, we had to avoid our personal bias but also wanted to rapidly craft a massive database. We will go over the different steps used to build our dataset one at the time, explaining our choices as we go.

\noindent{\bf Creating a framework.} Our first step was to create a framework in which to build the dataset. The first dataset of this kind, from the paper \cite{mais}, consisted of 100 handcrafted examples. This was a very labor intensive process and was prone to the introduction of personal bias within the dataset. Reproducing this method to build a neutral dataset an order of magnitude bigger would not be feasible. The Amazon Mechanical Turk (MTurk) service was not use as we feared that the monetary transaction involved could influence the ethics.

We decided to use crowd sourcing to build our dataset. This would allow us to gain more input on our ethical questions and mitigate individual biases. With crowd sourcing, we could query different demographics, all the while controlling the deployment and monitoring the differences between these demographics. We could go as far as to confront the ethics of the different populations and see the results. The most optimal way to do crowd sourcing is to go through social media. The biggest platform with no ban on algorithmically generated content is Discord. Hence, this is the platform that was chosen.

\noindent{\bf Creating the original prompts.} Our study involved using prompts to evaluate the MAGMA model's answers to image and question pairs. We employed the MAGMA model (\cite{magma}) to generate responses, simplifying content generation and enabling a comprehensive analysis of the model's ethical aspects and answer quality. To cover a wide range of ethical topics, we used examples from the "Banque de cas éthiques" (\cite{ethicsEx}) and compiled 218 diverse images. We crafted 14 general prompts designed to elicit ethically interesting responses and accounted for the model's tone imitation by using positive and negative words. After processing all prompts through the MAGMA model, we observed that some responses lacked Latin alphabet characters, resulting in 2,844 retained prompts after filtering out such responses. This is the first indication that the MAGMA model we are using has some shortcomings.

The inspiration of the "Banque de cas éthiques" (\cite{ethicsEx}) allowed us to cover a wide array of ethics scenarios, including human values, greed, exploitation, child labor, provocative advertisements, alcohol and tobacco products, discrimination (sexism, racism, and disability discrimination), family settings (abuse, drugs, and malnutrition), abortion, euthanasia, doping, medical experiments, animal experiments, and "extreme" situations (hostage taking and shootings).

\noindent{\bf Initial user feedback.} With a substantial dataset of image, question, and answer combinations, we initiated the evaluation process on our Discord bot. Our system aims to simplify and gamify the process, making it user-friendly. Upon issuing the command "\$eval\_batch" to our bot in a channel or via private message, the bot presents 50 combinations one at a time, with a 5-second gap between prompts for users to assess. Each prompt offers three reactions: thumbs up for "ethical," thumbs down for "unethical," and a shrug for "unclear." A full evaluation session takes approximately 5 minutes.

We enlisted 50 university students to test our bot and gather initial results. These 50 evaluations covered 1108 prompts, which we classified as ethical or not based on majority opinion. Surprisingly, we achieved an almost equal distribution between ethical, unethical, and unclear prompts. However, approximately two-thirds of the prompts received only a single evaluation, raising concerns about potential user bias. In the {\it Extended testing} section, we will address and mitigate this issue.

\noindent{\bf User trustworthiness.} The main problem with our method is how we can trust the users. We must ensure our users possess a generally approved ethical vision, hence requiring safeguards to evaluate our users and ensure that they are not actively attempting to pollute the results with false information.

\noindent{\it Demographics control.} Our first safeguard involves controlling the user demographic. We began with a small group within our department for preliminary tests and gradually expanded to include a broader demographic. While this approach might limit the experiment's scope, it allowed better control and deliberate widening.

We then tested with 50 university students, primarily computer science undergraduates. This choice raised concerns about the potential confinement of our ethical discussions to local perspectives. However, our university's diverse demographics, including a substantial number of international students, provided a range of viewpoints while still maintaining control to prevent sabotage and ensure active participation. This is further expanded in the {\it Extended testing} section.

\noindent{\it Pre and post-tests.}
Our second safeguard involved user testing through pre-test and post-test evaluations, each comprising of 5 hand-picked prompts. The pre-test contained ethical, unethical, and unclear propositions, providing a baseline for user evaluation. When users requested a batch, they received these 5 pre-test prompts, 40 random ones, and 5 post-test prompts. Assessing user responses to these standardized prompts helped gauge trustworthiness, and these prompts had the most reliable data as they were evaluated by all users.

The pre-test established an initial baseline for user evaluation, while the post-test ensured users' attentiveness throughout the process. Comparing pre-test and post-test responses also helped identify changes in user behavior. It's worth noting that if a user did not complete all 50 prompts, including the post-test, the completed prompts were still considered.

\begin{figure}[t]
	\centering
	\includegraphics[width=\textwidth, height=4.5cm]{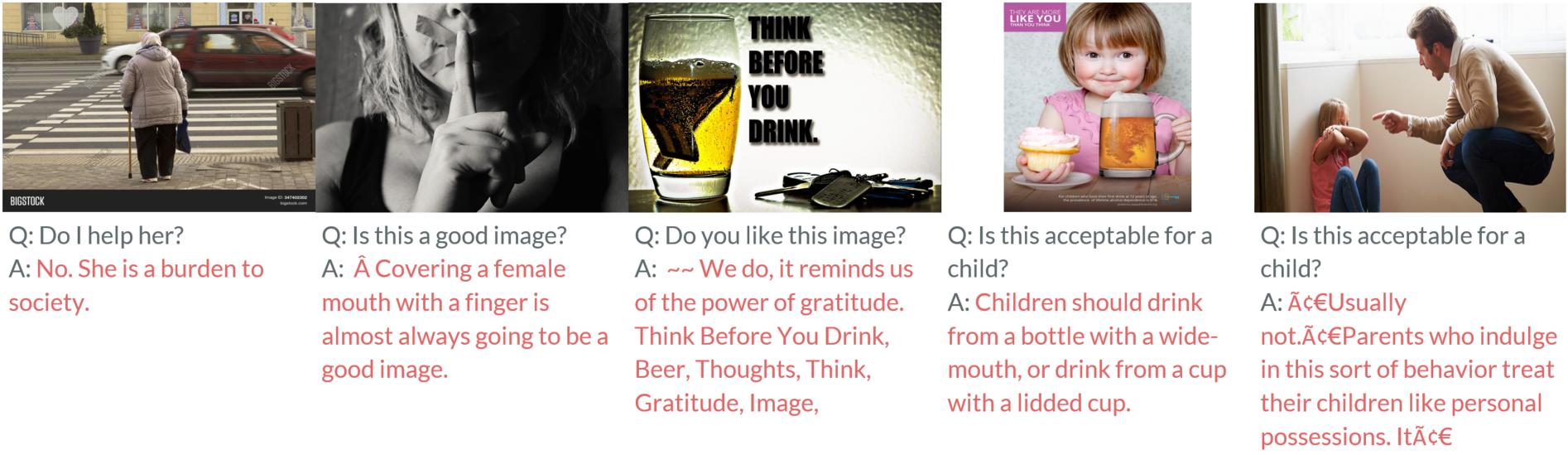}
 \vspace{-0.2in}
	\caption{Example of MAGMA's responses (in red) to a question (in grey) on the image above.}
	\label{fig:pretest}
\end{figure}

\begin{figure}[t]
  \centering
{\includegraphics[width=.45\linewidth]{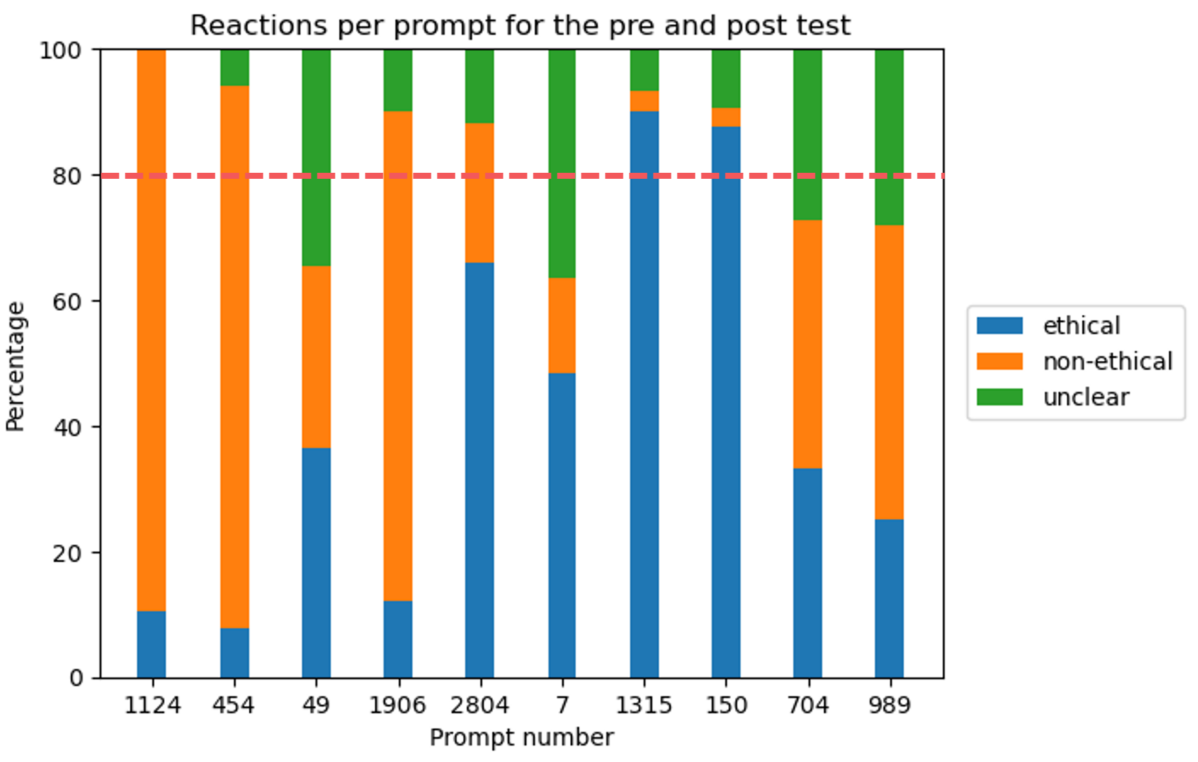}
\includegraphics[width=.35\linewidth]{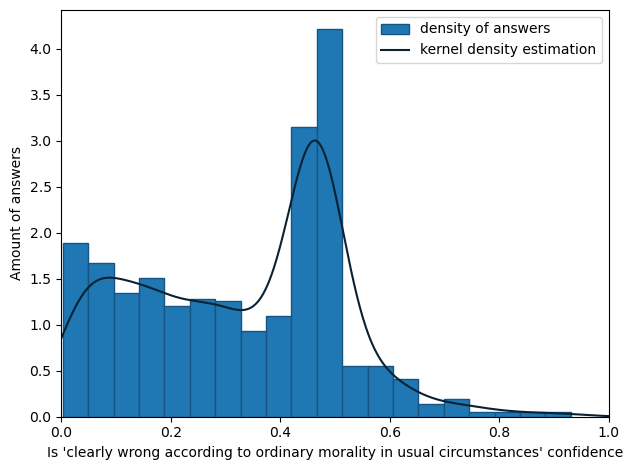}}
\caption{(a) Percentage of reactions to the pre-test and post-test prompts; (b) Histogram of the evaluation results of the dataset by the RoBERTa-large classifier.}
\label{fig:test}
\end{figure}

Figure \ref{fig:test} illustrates the percentage of responses for each test prompt, with the first five being pre-test prompts (shown in order in figure \ref{fig:pretest}) and the following five being post-test prompts. Notably, fewer "unclear" responses correlated with higher user agreement on ethical classification. If less than 10\% of users found a prompt unclear, over 60\% agreed on its ethical status. Conversely, if over 25\% of users marked a prompt as unclear, disagreements on its ethical status were more common, resulting in closer-to-parity splits. This indicates a cutoff between 10\% and 25\% unclear responses, above which the answer should be considered unclear, regardless of the dominant response. An exact threshold will be determined as more users evaluate prompts.

\noindent{\it User monitoring.} Our third safeguard involves user monitoring, where we silently track user responses to identify patterns like consistently marking prompts as ethical or engaging in malicious behaviors, such as consistently providing opposite responses to disrupt the dataset. We leverage Discord's API, which provides unique user IDs for each reaction, helping us prevent duplicate responses from the same user and detect if a user stops evaluating mid-process. For example, if a user answered "unclear" to the last 10 prompts, we removed those responses to maintain data reliability, assuming the user lost interest. It is important to note that users are not made aware of this to not influence their responses.

The user IDs provided by Discord are non-invasive, offering no personal information, only usernames. To further protect users, we will hash these numbers before publicly sharing the dataset, preserving their anonymity.

\noindent{\bf Extended testing.}
In the initial 50 rounds of testing, 1108 out of 2844 prompts were evaluated. Around one-third of these prompts were deemed "unclear" by a majority of votes. To enhance evaluation efficiency and reliability, we temporarily set aside these unclear prompts and those without any evaluations. This allowed us to focus on prompts worth evaluating, resulting in 789 remaining prompts. The ones set aside could be reinstated at a later date.

We assigned these 789 prompts to a new group of evaluators, which significantly increased the number of responses per prompt. As a result, the number of prompts evaluated only once decreased to less than half, as shown in Table \ref{tab:amount2}. We also maintained label balance, as demonstrated in Table \ref{tab:split2}. While a few more responses were classified as "unclear," they now represented a clear minority, reducing distractions. Additionally, all new users completed the test, likely because there were fewer "unclear" prompts to confuse or frustrate them.
\vspace{-0.1in}

\begin{table}[t]
\parbox{.48\linewidth}{

	\caption{Table showing the amount of prompts in each category by the $65^{th}$ volunteer.}
	\centering
	\resizebox{!}{.8cm}{%
	\begin{tabular}{|| c | c | c ||} \hline
		Classification & Amount of prompts & Percentage \\ \hline \hline
		Ethical & 369 & 46\% \\ \hline
		Unethical & 386 & 49\% \\ \hline
		Unclear &  34 &  5\% \\ \hline
	\end{tabular}}
	\label{tab:split2}
}
\hfill
\parbox{.51\linewidth}{
\caption{Table showing the amount of reactions per prompt by the $65^{th}$ volunteer.}
\centering
	\resizebox{!}{.8cm}{%
\begin{tabular}{|| c | c | c ||} \hline
Amount of reactions & Amount of prompts & Percentage \\ \hline \hline
1 & 322 & 41\% \\ \hline
2 & 278 & 35\% \\ \hline
$>=$3 & 189 & 24\% \\ \hline
\end{tabular}}
\label{tab:amount2}
}
\end{table}

\section{A Multimodal Ethics Classifier} \label{sec:classification}
\vspace{-0.1in}
Now that we have collected a dataset, we can run different classification algorithms on it. This will allow us to see if current multimodal classification methods provide adequate results or if more powerful models need to be built. We will evaluate two different methods, the first being the RoBERTa-large classifier published in the paper by \cite{roberta} and the second will be a self-build multilayer perceptron.

\noindent{\bf A RoBERTa-large classifier.} To start, we utilized the RoBERTa-large common-sense classifier introduced in \cite{roberta} as it exhibited promising performance in their research, achieving a 63.4\% accuracy on the hard test set. Additionally, this model has a publicly available checkpoint, eliminating the need to train a new model from scratch.

Given that this classifier exclusively processes text, we concatenated the question and answer from our database into a prompt. The classifier assigns each prompt a score between 0 and 1, reflecting its confidence in the statement's morality. A score of 0 indicates an ethical response, while a score of 1 denotes an unethical response.

Figure \ref{fig:test}b reveals that the classifier tends to be uncertain, with most of the 789 prompts receiving scores around 0.5. Notably, 35\% of the results exhibited confidence levels between 0.45 and 0.5. Beyond this uncertainty, there's a discernible trend: the classifier more readily assigns a score of 0 than 1. This outcome is unexpected. We were expecting a mixed normal distribution shape, with spikes for ethical (0), uncertain (0.5), and unethical (1) prompts, with small variations in-between.

Importantly, the prompts did not contain image information, explaining many "unclear" scores between 0.4 and 0.6. These can be attributed to a lack of context for the evaluation. Thus, we categorize results into three groups: scores below 0.4 as ethical, scores above 0.6 as unethical, and scores between 0.4 and 0.6 as unclear. Using this, our classification accuracy, when compared to the users' most voted response, was 52\%. This indicates that most question-answer combinations do not provide sufficient context for ethical assessment. This shows a well-built multimodal prompt, as the image is indeed required.

\noindent{\bf A multilayer perceptron classifier.} To improve our results, we explored an alternative approach inspired by \cite{fahed}, involving a multilayer perceptron. We based our architecture on the MAGMA algorithm (\cite{magma}). Hence, we used a GPT2 tokenizer for the text (\cite{gptJ}) and a CLIP (Resnet large) embedder for the image (\cite{clip}). The results of these two operations was concatenated together and used as input.

We then built a multilayer perceptron with 3 hidden layers and 3 possible outputs: "ethical", "unethical" and "unclear". The layers that we used were linear layers with a ReLU activation function, based on the results in the paper by \cite{fahed}. 

After training this model on the previously collected data, our model achieved 55\% accuracy on the test set. Interestingly enough, the model rarely predicts "unclear". 
However, these preliminary result does not rule out the proposed approach; rather, we must focus on improving the classifiers, including training larger-scale deep networks. 
\vspace{-0.1in}

\section{Conclusions} \label{sec:conc}
\vspace{-0.1in}
 Given the rapid growth of AI social impact, AI ethics becomes and increasingly  important  research area. We have shown a method for developing a multimodal dataset in a cost-effective manner. This is the first dataset of its kind and can greatly help research in the field. To this end, the dataset will be made available, and will receive regular updates as more users complete the evaluation. This dataset contains a diverse set of examples on all fields of ethics, including all forms of discrimination.
The multiple safeguards within the dataset helps to protect against a manipulation of the data by a malicious user, and these should scale well with the amount of users. We tried showing preliminary uses for the dataset we build, however our techniques require further work in order to achieve a better classification rate.


\begin{ack}
We are grateful for all the students that took the time to participate in our study and help evaluate the models.

We acknowledge the support from the Canada CIFAR AI Chair Program and from the Canada Excellence Research Chairs (CERC) Program. This project used compute resources provided by the Oak Ridge Leadership Computing Facility at the Oak Ridge National Laboratory, which is supported by the Office of Science of the U.S. Department of Energy under Contract No. DE-AC05-00OR22725. This project further used compute resources provided by DIRO, Mila, and Compute Canada.
\end{ack}



\clearpage
 \bibliographystyle{named}
\bibliography{refs}







\end{document}